\documentclass[runningheads]{llncs}
\usepackage{amsmath}
\usepackage{amsfonts}
\usepackage{graphicx}
\usepackage{hyperref}
\usepackage[T1]{fontenc}
\usepackage[utf8]{inputenc}
\usepackage[english]{babel}
\usepackage{booktabs}
\usepackage[table,xcdraw]{xcolor}
\usepackage{tabularx} 
\usepackage[table,xcdraw]{xcolor}
\usepackage{todonotes}
\definecolor{contentclr}{rgb}{0.1,0.3125,0.54}
\definecolor{urlclr}{rgb}{0,0.254,0.426}
\definecolor{citeclr}{rgb}{0.125,0.414,0.363}
\hypersetup{
    colorlinks=true,
    linkcolor=contentclr,
    urlcolor=urlclr,
    citecolor=citeclr
}

\rowcolors{3}{gray!15}{white}

\begin{document}
\title{Conformal Adversarial Generative Ensemble}
%
%
\author{Ahmad Shahi\inst{1,*,\dag} \and
Mamehgol Yousefi\inst{2, *} \and
Brendon J.~Woodford\inst{3} \\
Farhaan Mirza\inst{4} \and
Tapabrata Chakraborti\inst{5}}
\authorrunning{A. Shahi et al.}
%
\institute{
Unitec Institute of Technology, Auckland, New Zealand \\
\email{ahmad.shahi@gmail.com} \and
School of Product Design, University of Canterbury, Christchurch, New Zealand \\
\email{mahgol.yousefidashliboroun@pg.canterbury.ac.nz}  \and
School of Computing, University of Otago, Dunedin, New Zealand \\
\email{brendon.woodford@otago.ac.nz} \and
Auckland University of Technology, Auckland, New Zealand \\
\email{farhaan.mirza@aut.ac.nz} \and
The Alan Turing Institute, University College London, United Kingdom \\
\email{t.chakraborty@ucl.ac.uk; tchakraborty@turing.ac.uk}
}

\maketitle              
\noindent\hspace{2.5em}* Both are joint first authors. 
\textsuperscript{\dag} Corresponding author.
\begin{abstract}
Accurate time series forecasting is critical across various domains, yet traditional ensemble methods often suffer from the disproportionate influence of extreme forecasts. We introduce the Conformal Adversarial Generative Ensemble (CAGE), a novel framework that combines generative modeling, adversarial discrimination, and conformal prediction to enhance forecast reliability and accuracy. CAGE employs multiple generative models to produce initial forecasts, which are then evaluated by a discriminative component using conformal prediction techniques. \textit{p}-values derived from nonconformity scores help dynamically adjust model weights, minimizing the impact of unreliable forecasts. This approach ensures that only the most credible predictions contribute to the final ensemble output. Our empirical and statistical analyses of time series data from New Zealand's milk collection and the global health data from the public owid-monkeypox dataset show that the CAGE outperforms traditional ensemble methods, especially in handling outliers and noisy data. 
By incorporating conformal prediction, CAGE delivers accurate and statistically rigorous forecasts, enhancing decision-making. We have demonstrated performance on two different datasets deliberately to showcase that the proposed method offers a versatile solution potentially applicable across finance, weather, and supply chain management.

\end{abstract}
\section{Introduction}

Time series forecasting is a fundamental task in various fields, including finance, weather prediction, healthcare, supply chain management, and energy consumption forecasting. Accurate forecasting models are crucial for informed decision-making and strategic planning. Traditional ensemble methods, which combine multiple predictive models to enhance accuracy, have been widely used due to their ability to mitigate individual model weaknesses and improve overall performance \cite{dietterich2000ensemble,breiman1996bagging}. 

Simple averaging of base learners remains a popular method due to its simplicity and effectiveness, particularly in reducing variance. For weather forecasting, \cite{duan2021comparing} applied simple averaging techniques to climate modelling, showing enhanced stability in predictions across different climate scenarios \cite{duan2021comparing}. Similarly,\cite{Sloughter2007ProbabilisticQP} explored this technique in the same context and found that simple averaging, despite its simplicity, was remarkably effective at improving forecast reliability over individual models \cite{Sloughter2007ProbabilisticQP}. This method has been applied in various domains such as healthcare. Yoon et al. \cite{yoon2023multi} implemented a multi-model ensemble to forecast the diagnosis of cardiovascular diseases. They compared the proposed method with simple average ensemble. Their findings indicate that even straightforward ensemble techniques can effectively capture the dynamics of disease spread, improving the reliability of public health responses \cite{yoon2023multi}.

The weighted ensemble method, where models are assigned importance based on their performance, has seen considerable interest. Cheng et al. \cite{cheng2012comparison} demonstrated the efficacy of a weighted ensemble approach in financial forecasting, where each model's prediction is weighted according to its historical accuracy, significantly outperforming standard averaging methods. In the same domain, Adhikari et al. \cite{adhikari2016time} utilised a weighted ensemble approach that dynamically adjusts weights based on real-time performance, improving prediction accuracy by adapting to market volatility and capturing non-linear patterns inherent in the data. This method has been used in various domains such as health-related predictions. Dutta et al. \cite{dutta2022early} employed a weighted ensemble of classifiers, including Naive Bayes, Random Forest, Decision Tree, XGBoost, and LightGBM for predicting diabetes early using a newly labeled dataset from Bangladesh. By dynamically adjusting model contributions, the method effectively handles data complexities and outperforms traditional single-model approaches \cite{dutta2022early}.

A sophisticated form of ensemble where predictions from various models are used as inputs to a second-level model has been extensively studied. Bhagat et al. \cite{bhagat2024efficient} applied stacking techniques to early heart attack prediction, using outputs from several machine learning methods as inputs to a meta-learner with Logistic Regression or another generalized linear model. This method not only enhanced accuracy but also adapted more dynamically based on real-time performance, improving the model's ability to handle complexities in prediction data. However, this approach involves challenges such as the complexity of integrating multiple models and the potential for overfitting, especially when the meta-learner is overly tuned to specific training datasets \cite{wolpert1992stacked}. Additionally, the computational demand and expertise required for optimally configuring multiple layers of models can be significant, potentially limiting real-time applications due to resource constraints \cite{gupta2022stacking,jakhar2024self}.

Moreover, the integration of deep learning with traditional ensemble techniques has furthered the development of robust forecasting models. He et al. \cite{He2023FinancialTS} developed a deep ensemble model combining LSTM, GRU, and CNN to improve financial time series forecasting accuracy. Their approach captures non-linear patterns in financial data and outperforms traditional methods, despite requiring significant computational resources \cite{He2023FinancialTS}. Similarly, Tang et al. \cite{Tang2023} developed a deep ensemble framework for medical time series data, integrating CNNs and GNNs to improve classification and prediction tasks such as atrial fibrillation recurrence and seizure detection \cite{Tang2023}.
Deep ensembles are known for their robustness in capturing complex patterns. However, they face several drawbacks, such as high computational costs \cite{gupta2022stacking} and extensive data requirements to train multiple deep models effectively \cite{ganaie2022ensemble}. Moreover, their complexity can lead to difficulties in model interpretation and integration into real-time systems due to latency issues \cite{hu2021model}. These challenges highlight the necessity for ongoing innovation and refinement within the field.

In conclusion, the field of ensemble methods for time series forecasting is rapidly evolving, with significant contributions that leverage both traditional and innovative techniques. The continuous refinement of these methods promises to address the complex demands of modern forecasting tasks effectively.
However, due to the complexity and computational costs associated with deep ensembles and model stacking, as mentioned earlier, our focus remains on traditional ensemble methods. Despite their utility, these methods also present some drawbacks, which are summarized below:

\begin{itemize}
    \item Exhibits significant limitations for instance in simple averaging, such as equal weighting of potentially diverse quality models, leading to suboptimal performance \cite{Yang2002Multistage}, and sensitivity to biased models which can skew overall predictions \cite{brown2005managing}.
    \item Additionally, its inability to adapt to evolving data patterns restricts its applicability in dynamic environments \cite{hammami2020neural}.
    \item Although the weighted ensemble method offers improved performance by assigning different weights to models based on their accuracy, it also faces challenges such as the complexity of determining optimal weights \cite{rokach2010ensemble} and potential overfitting if model weights are excessively tuned to specific data sets \cite{kuncheva2014combining}.
    \item They struggle with the disproportionate influence of extreme forecasts, particularly in the presence of noisy or outlier data \cite{krawczyk2016learning,liu2024ensemble}. 
    
\end{itemize}

To address these challenges, we propose the Conformal Adversarial Generative Ensemble (CAGE), a novel framework that synergises generative modeling, adversarial discrimination, and conformal prediction. The CAGE framework leverages the strengths of each component to produce robust and reliable forecasts. Generative models within the ensemble generate initial forecasts, which are then evaluated by a discriminative component using conformal prediction techniques. This evaluation assigns \textit{p}-values based on nonconformity scores from a calibration set, enabling dynamic adjustment of model weights and minimising the impact of unreliable forecasts.

Conformal prediction, a statistical technique that provides valid measures of uncertainty, plays a crucial role in the CAGE framework. It ensures that the assigned \textit{p}-values reflect the plausibility of each forecast, enhancing the reliability of predictions and providing a quantifiable measure of uncertainty crucial for decision-making processes \cite{vovk2005algorithmic,shafer2008tutorial}.
The ability of the CAGE framework to dynamically adjust model weights based on forecast reliability sets it apart from traditional ensemble methods, offering significant advantages in handling outliers and noisy data. By ensuring that only the most credible forecasts influence the final output, CAGE improves both the accuracy and robustness of predictions.

Empirical evaluations on two very distint predictive tasks using benchmark datasets (NZ milk collections and global monkeypox tracking) demonstrate that CAGE outperforms traditional ensemble methods, providing more accurate forecasts and effectively handling extreme values and noise. This underscores its utility for complex and dynamic forecasting tasks.

In summary, CAGE represents a significant advancement in ensemble forecasting methodologies. This paper details CAGE’s architecture, its theoretical underpinnings, and empirical performance, highlighting its potential for wide-ranging applications.

\section{Inherent Challenges of Existing Ensemble Methods}

\begin{figure}[ht]
\centering
\includegraphics[scale=0.5]{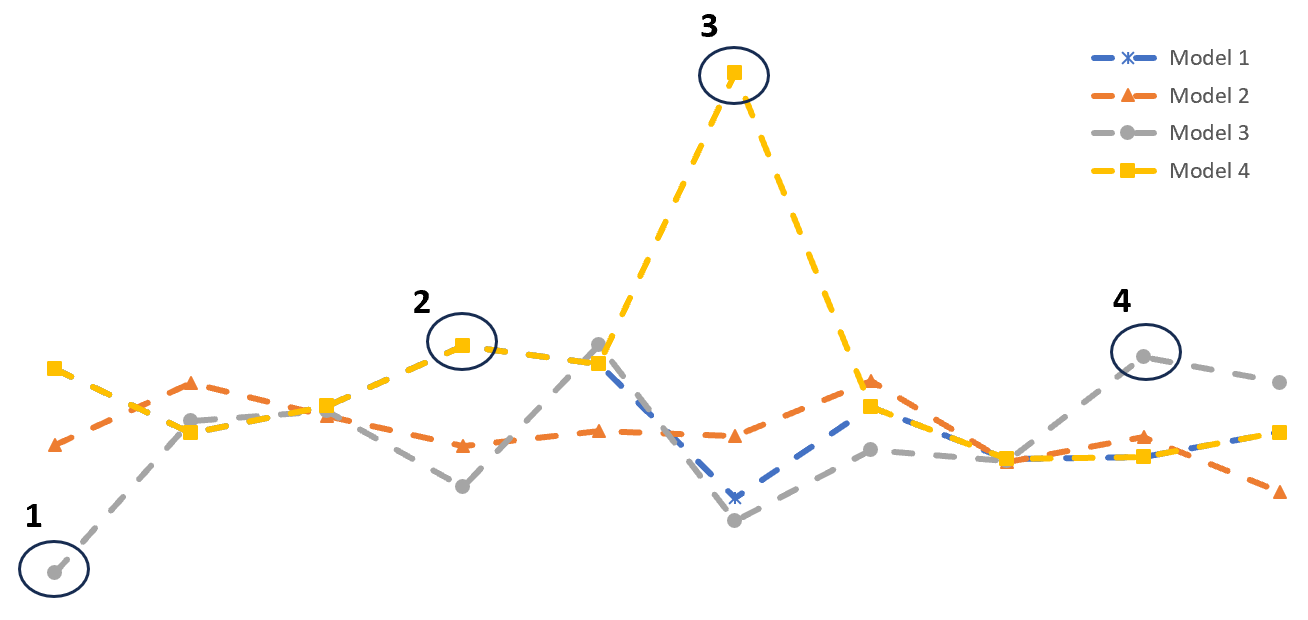}
\caption{Forecast examples from four different models. Points 1-4 are potential outliers or invalid values.}
\label{fig: Challenges}
\end{figure}

\begin{figure}[h]
\centering
\includegraphics[scale=0.5]{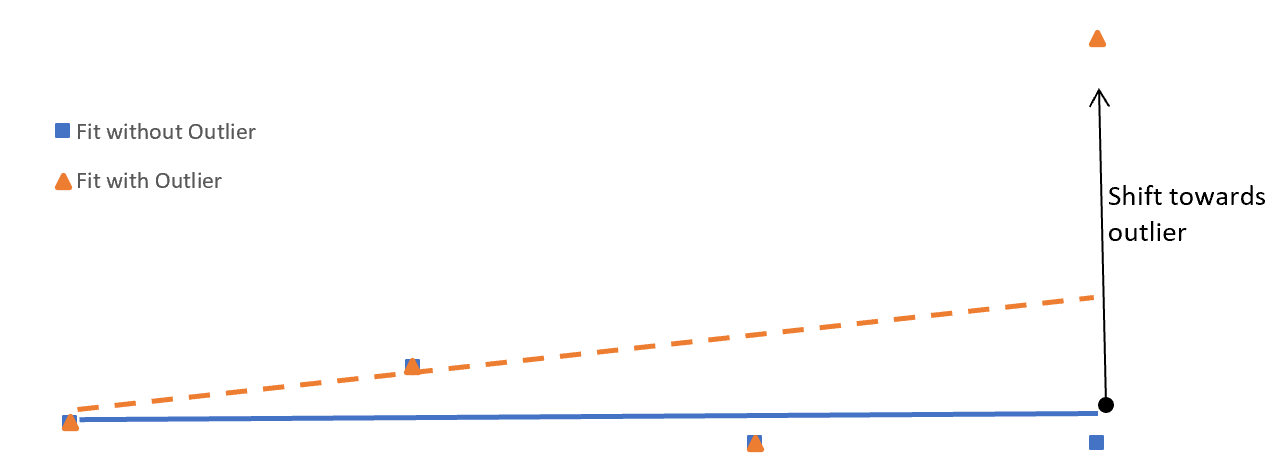}
\caption{Impact of outliers on a model's fit.}
\label{fig:lr_outlier}
\end{figure}

\begin{figure}[h]
\centering
\includegraphics[scale=0.5]{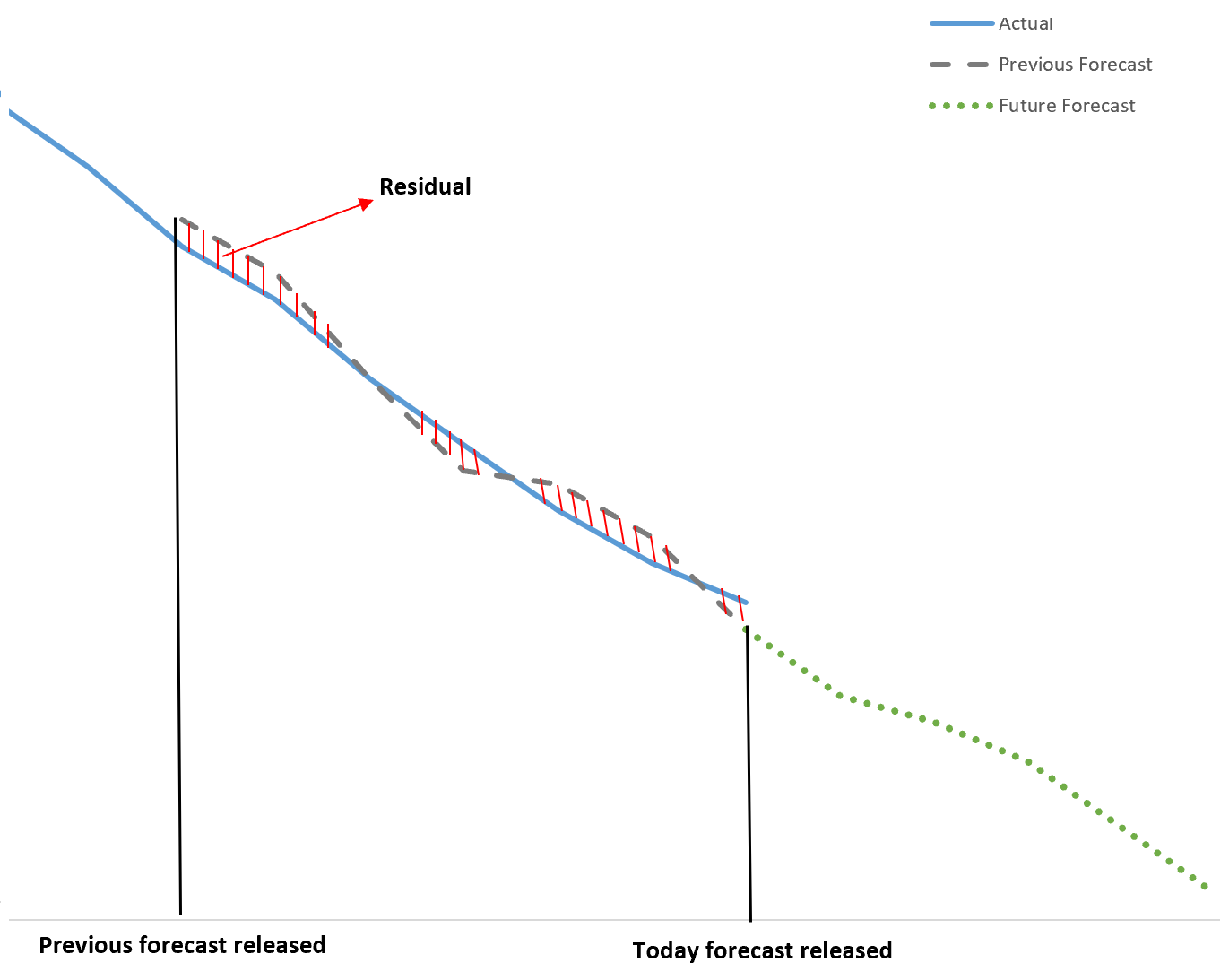}
\caption{Demonstration of model bias based on previous performance, highlighting the need for adaptive bias correction.}
\label{fig:bias_example}
\end{figure}

Traditional ensemble methods have been pivotal in advancing the accuracy of time series forecasting by capitalising on the diverse strengths of multiple predictive models. However, significant challenges persist that can compromise their effectiveness in real-world scenarios, as illustrated in Figures \ref{fig: Challenges}, \ref{fig:lr_outlier}, and \ref{fig:bias_example}.
Figure~\ref{fig: Challenges} exemplifies forecasts from four different models, demonstrating how potential outliers or invalid values (Points 1-4) challenge decision-making: should these points be included or excluded in the ensemble calculation? The dilemma underscores the need for robust validation methods before data aggregation.
Figure~\ref{fig:lr_outlier} exemplifies how outliers can significantly affect the fitting of a model, potentially leading to erroneous predictions and highlighting the necessity for effective outlier detection and handling mechanisms.

Traditional models often fail to capture complex patterns and interdependencies, as shown in Figure \ref{fig:bias_example}, resulting in residual errors and unreliable forecasts. To address these limitations, it is crucial to implement methods that dynamically adjust model biases based on recent performance to enhance forecast accuracy, particularly in the short term. The ensemble methods inherently possess issues highlighted in earlier Figures \ref{fig: Challenges} - \ref{fig:bias_example}. These methods, while combining the strengths of multiple models, also compound their weaknesses. We summarise the inherent challenges as follows:

\textbf{Disproportionate Influence of Extreme Forecasts:} Ensemble methods such as simple and weighted averages are vulnerable to the disproportionate influence of extreme forecasts \cite{krawczyk2016learning,liu2024ensemble}. Highly deviant predictions from individual models can skew the aggregated outcome, especially during atypical market conditions.

\textbf{Difficulty in Handling Noisy Data:} Noisy data complicates the distinction between transient changes and lasting trends, leading to less reliable forecasts and diminished model performance over time.

\textbf{Lack of Robust Uncertainty Quantification:} Traditional ensemble methods typically lack a robust mechanism for quantifying the uncertainty of their predictions, a critical drawback in risk-sensitive decision-making processes.

This section highlighted the inherent challenges in existing ensemble methods, emphasizing the need for enhanced techniques that can better manage outliers, noise, and prediction uncertainties.
To address these challenges, we propose the Conformal Adversarial Generative Ensemble (CAGE), which integrates generative modelling, adversarial discrimination, and conformal prediction. This proposed framework is designed to mitigate the issues of extreme forecasts, handle noisy data effectively, provide quantifiable measures of forecast uncertainty, and enhance computational efficiency, making it suitable for a wide range of forecasting applications.

\section{Conformal Adversarial Generative Ensemble (CAGE)}
\label{cage}
The Conformal Adversarial Generative Ensemble (CAGE) is designed to enhance time series forecasting by combining the strengths of generative modelling, adversarial discrimination, and conformal prediction. This section details the theoretical underpinnings and operational framework of CAGE.

\begin{figure}[htbp]
\centering
\includegraphics[width=\linewidth]{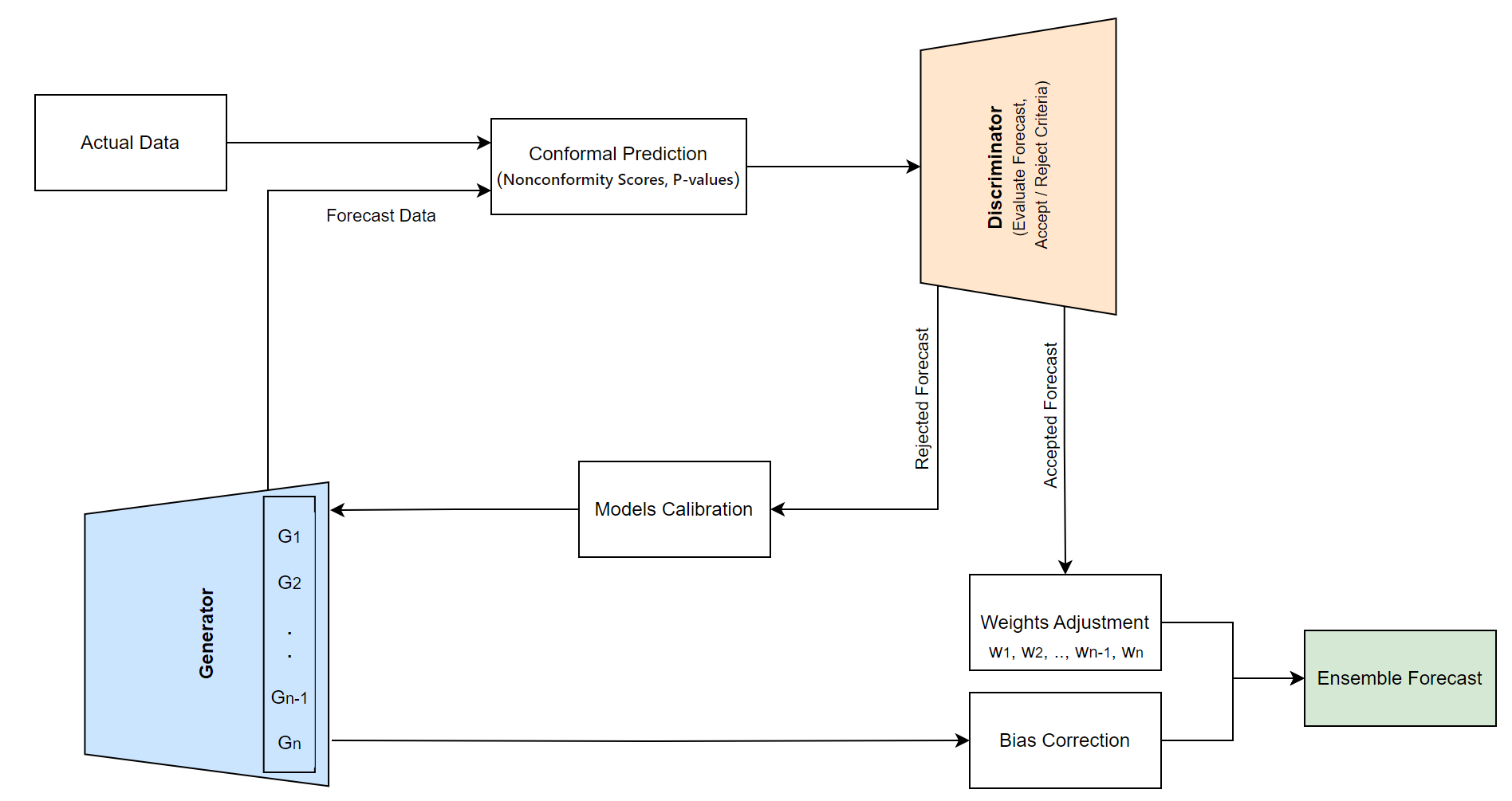}
\caption{Diagram of the Conformal Adversarial Generative Ensemble (CAGE) framework}
\label{fig:cage_framework}
\end{figure}

\textbf{Generative Models (G):}
Generative models are used to produce initial forecasts for the given time series data. These generative models can be of various types, such as statistical-based learning models like exponential smoothing, or machine learning-based models like decision trees. The diversity of models ensures that different aspects and patterns of the time series data are captured. Therefore, let \( G_i \) represent the \(i\)-th generative model in the ensemble. For an input \( x \), the generative model \( G_i \) produces a forecast \( \hat{y}_i \):

\begin{equation}
    \hat{y}_i = G_i(x)
\end{equation}

In this study, the generative models we used are Ridge regression, Elastic-Net, RANdom SAmple Consensus (RANSAC), random forest, XGBoost, light Gradient Boosting Machines (lightGBM), catboost, and K-Nearest Neighbors (KNN) to capture linear and non-linear relationship from historical data.

\textbf{Conformal Prediction (C):}
Conformal prediction provides a framework for assigning confidence measures to the forecasts generated by the generative models. This is achieved through the calculation of nonconformity scores and \textit{p}-values.

\paragraph{Nonconformity Score Calculation:}
For each observation \( (x_j, y_j) \) in the calibration set, generate predictions using the trained models and calculate the nonconformity score:

\begin{equation}
    \hat{y}_{i,j} = G_i(x_j)
    \label{eq:prediction}
\end{equation}

\begin{equation}
    \alpha_{i,j} = |y_j - \hat{y}_{i,j}|
    \label{eq:nonconformity}
\end{equation}

In the equations provided, Eq. \ref{eq:prediction} calculates the predicted value (\(\hat{y}_{i,j}\)) for the \(j\)-th observation using the \(i\)-th model, where \(G_i\) represents the generative function of the \(i\)-th model, and \(x_j\) denotes the features of the \(j\)-th observation. The Eq. \ref{eq:nonconformity}, defines the nonconformity score (\(\alpha_{i,j}\)) for the \(j\)-th observation as calculated by the \(i\)-th model, quantifying the absolute difference between the actual observed value \(y_j\) and the predicted value \(\hat{y}_{i,j}\). Here, \(i\) indexes the specific model being used, and \(j\) indexes the observations within the dataset, indicating that each observation is processed through each model to assess prediction accuracy.

\paragraph{Nonconformity Distribution:}
Aggregate the nonconformity scores for each model across the calibration set:

\begin{equation}
    \text{Nonconformity Distribution}_i = \{\alpha_{i,1}, \alpha_{i,2}, \ldots, \alpha_{i,n}\}
\end{equation}

\paragraph{\textit{p}-value Calculation:}
For a new prediction \( \hat{y}_i \) corresponding to input \( x \), calculate the nonconformity score \( \alpha_i = |y_{\text{true}} - \hat{y}_i|\).
\noindent The \textit{p}-value Compute  based on the nonconformity distribution:

\begin{equation}
    p_i = \frac{|\{ j : \alpha_{i,j} \geq \alpha_i \}| + 1}{n + 1}
\end{equation}

This \textit{p}-value provides a measure of how unusual the new forecast's error is compared to the historical errors in the calibration set. \( n \) represents the total number of observations.

\textbf{Discriminator Feedback (D):}
The discriminator evaluates each forecast using the \textit{p}-value and a predefined threshold \( \alpha \). In the discriminator, the acceptance and rejection criteria are defined as follows:
\begin{itemize}
    \item \textbf{Acceptance Criterion}: A forecast is accepted if its \textit{p}-value \( p_i \geq \alpha \). 
    \item \textbf{Rejection Criterion}: A forecast is rejected if its \textit{p}-value \( p_i < \alpha \).
\end{itemize}
Accepted forecasts, considered reliable, directly contribute as base learners in the ensemble, influencing the final decision-making. Rejected forecasts are calibrated and tuned for accuracy before potential reintegration. This method ensures that only the most credible forecasts impact the ensemble’s output, enhancing prediction reliability and accuracy.

\paragraph{Adjusting Weights:}
For each forecast, dynamically adjust the weights based on acceptance or rejection. If a forecast is rejected, set its weight to zero or reduce it significantly to minimize its impact on the ensemble:

\begin{equation}
    w_i = 
    \begin{cases} 
    \frac{p_i}{\sigma_i + \epsilon} & \text{if } p_i \geq \alpha \\
    0 & \text{if } p_i < \alpha 
    \end{cases}
\end{equation}

In this framework, \( \sigma_i \) typically denotes the measure of variability or uncertainty for the \( i \)-th forecast, such as standard deviation, while \( \epsilon \) is a small positive constant added to stabilize the calculation and prevent division by zero. The weight \( w_i \) for each forecast is dynamically adjusted depending on its \( p \)-value \( p_i \) compared to a threshold \( \alpha \). If \( p_i \geq \alpha \), the weight is calculated as \( \frac{p_i}{\sigma_i + \epsilon} \), combining the forecast's relevance and its variability. If \( p_i < \alpha \), the weight is set to zero, effectively excluding it from influencing the ensemble. The rejected forecasts are calibrated and tuned for accuracy before potential reintegration.

After adjusting the weights of individual models, each weight is normalised by dividing it by the total sum of all weights of the generative models. This ensures that the combined weight of all models equals one.

\begin{equation}
    w_i = \frac{w_i}{\sum_{k=1}^{N} w_k}
\end{equation}

where \( N \) is the number of generative models.

\textbf{Bias Correction:} Bias is adjusted for each model based on its recent performance.

\begin{equation}
    \text{Bias}_i = \text{mean}(y_{\text{true}} - \hat{y}_i)
    \label{bias}
\end{equation}

\begin{equation}
    \hat{y}_{\text{corrected}, i} = \hat{y}_i - \text{Bias}_i
    \label{sub_bias}
\end{equation}

The bias for each model \( i \), denoted \( \text{Bias}_i \), is calculated as the mean difference between the true values \( y_{\text{true}} \) and the model's predictions \( \hat{y}_i \) (Eq. \ref{bias})). The corrected predictions \( \hat{y}_{\text{corrected}, i} \) are then obtained by subtracting this bias from the original predictions (Eq. \ref{sub_bias}).

\textbf{Ensemble Forecast:} The ensemble method aggregates the corrected forecasts as follows:

\begin{equation}
    \hat{y}_{\text{ensemble}} = \sum_{i} w_i \cdot \hat{y}_{\text{corrected}, i}
    \label{eq: en}
\end{equation}

As illustrated in Eq. \ref{eq: en}, the corrected predictions \(\hat{y}_{\text{corrected}, i}\) from each model are weighted by their respective normalized weights \(w_i\). This operation computes a weighted average forecast, \(\hat{y}_{\text{ensemble}}\), which is the sum of the weighted corrected predictions from all models indexed by \( i \). This method enhances the overall forecast's accuracy by integrating the individual strengths of each model.

\section{Results and Discussions}
\setcounter{footnote}{0}
In this study, we developed the CAGE ensemble method and conducted evaluations of it using time series data on milk collection across various geographical regions in New Zealand. Each region exhibits unique distribution patterns and volatility levels. For confidentiality purposes, the names of the regions have been anonymised and replaced with arbitrary labels throughout this paper. 
Additionally, we utilised the public time series OWID-monkeypox dataset\footnote{Data sourced from the OWID-monkeypox dataset available on Kaggle: \url{https://www.kaggle.com/datasets/utkarshx27/mpox-monkeypox-data}} for six different countries to validate the proposed method. For evaluation, we have adopted the Mean Absolute Error Percentage (MAE\%) as a key performance indicator to evaluate the accuracy of our forecasting models, valued for its clarity and interpretability \cite{vandeput2023demand}.

\subsection{Mean Absolute Error Percentage (MAE\%)}
\label{mape_percentage}
Mean Absolute Error Percentage (MAE\%), as proposed by \cite{vandeput2023demand}, is a scaled version of the Mean Absolute Error (MAE) that offers a more interpretable measure of forecast accuracy by accounting for the scale of the data. MAE alone does not indicate the adequacy of forecast accuracy without contextual scale; for instance, an MAE of 10 units is excellent if the average demand is 1000 units but poor if it's only 1 unit. MAE\% addresses this by dividing the MAE by the average actual demand and expressing it as a percentage:

\begin{equation}
\text{MAE\%} = \frac{\sum |a_j - p_j|}{\sum a} \times 100
\end{equation}

where:
\begin{itemize}
  \item \( a_j \) represents the actual value at index \( j \),
  \item \( p_j \) represents the predicted value at index \( j \),
  \item \( e_j = a_j - p_j \) is the error at index \( j \), which is the difference between the actual and predicted values,
  \item \( |a_j - p_j| \) is the absolute error at index \( j \), ensuring all errors are positive
\end{itemize}

MAE\% provides a normalized, interpretable metric that contextualizes the error relative to the scale of actual values, offering a robust Key Performance Indicator (KPI) for evaluating prediction models across different datasets or models. Unlike Mean Absolute Percentage Error (MAPE), MAE\% aggregates the total errors rather than calculating for each period, enhancing its simplicity and effectiveness in model assessment.

\subsection{Experimental results}
In this study, we utilised two datasets: 1) Dairy farming milk collection data from six regions in New Zealand, covering June 2007 to June 2024. This dataset records the date of collection, the region, and the volume of milk collected in litres per day. 2) The public owid-monkeypox dataset. Although it offers several attributes, our analysis focuses primarily on total monkeypox cases across different locations. It spans May 2022 to May 2023 with daily updates, and for our research, we have chosen data from six countries from different continents. 
We developed several machine learning and statistical learning methods as mentioned in Section \ref{cage} as base learners for forecasting milk collection 365 days ahead, assisting stakeholders in immediate and tactical planning and decision-making. For the second dataset, we forecast total cases 30 days ahead. We compared the proposed CAGE method with other ensemble methods, specifically  Simple Average ensemble (SimpleAvg) and Weighted Ensemble (wEnsemble) which served as baseline methods for comparison. 
We developed machine learning and statistical models, as detailed in Section \ref{cage}, to forecast milk collection 365 days ahead, aiding stakeholders in immediate and tactical planning and decision-making. For the second dataset, we predicted total cases 30 days ahead. We evaluated our CAGE method against baseline ensemble methods, namely Simple Average (SimpleAvg) and Weighted Ensemble (wEnsemble).
The simple average ensemble takes the average of forecasts produced by each generative model in the ensemble. This method assumes equal importance for all models. Let \( \hat{y}_{\text{s}} \) represent the forecast from the simple average ensemble for input \( x \):

\begin{equation}
    \hat{y_{\text{s}}} = \frac{1}{N} \sum_{i=1}^N \hat{y}_i = \frac{1}{N} \sum_{i=1}^N G_i(x)
\end{equation}

where \( N \) is the number of generative models in the ensemble.
The weighted ensemble assigns a weight \( w_i \) to each model’s forecast, reflecting the relative importance or performance of each model. Let \( \hat{y}_{\text{w}} \) represent the forecast from the weighted ensemble for input \( x \):

\begin{equation}
    \hat{y}_{\text{w}} = \sum_{i=1}^N w_i \hat{y}_i = \sum_{i=1}^N w_i G_i(x)
\end{equation}

where, \( w_i \) are the weights assigned to each model such that \( \sum_{i=1}^n w_i = 1 \).
The model stacking ensemble although computationally high \cite{gupta2022stacking} exhibited results nearly identical to those of the weighted ensemble. Given their similarity and conciseness, we opted not to detail the model stacking ensemble analysis, focusing instead on more distinct comparative insights. 

\begin{figure}[h]
\centering
\includegraphics[scale=0.24]{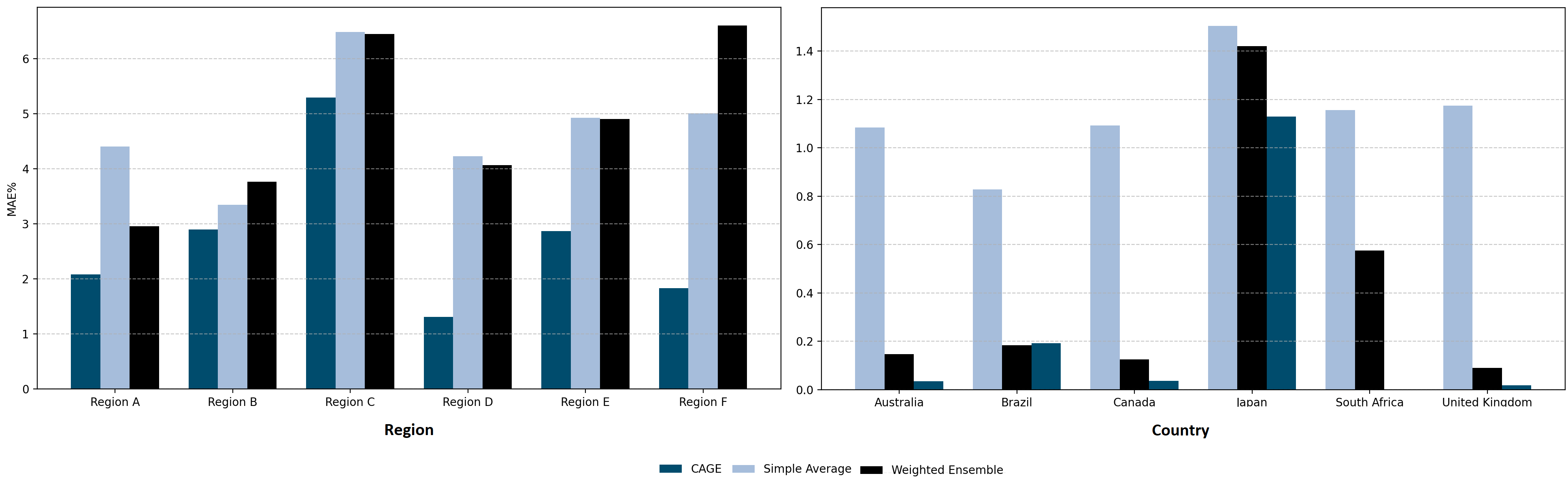} 
\caption{Comparison of MAE\%: The left panel shows results for milk collection data, and the right panel presents results for OWID-monkeypox data using CAGE versus baseline models.}
\label{fig:mape_region}
\end{figure}

For comparative analysis, as elaborated in Section \ref{mape_percentage}, we used the MAE\%. The analysis of MAE\% across different regions (milk collection data) and countries (owid-monkeypox data) provides insight into the predictive performance of each method in various geographical areas. Figure \ref{fig:mape_region} summarises the MAE\% for each method within each region and country. The detailed numbers are illustrated in Table \ref{table:mape} which presents the mean MAE\% and standard deviation ($\pm$ std) for each ensemble method across different regions and countries. This provides an overview of the performance and variability of each method.

\begin{table}[h]
\centering
\caption{MAE\% (mean $\pm$ std) Across Various Regions using CAGE vs Baselines}
\label{table:mape}
\begin{minipage}{0.49\textwidth}
\centering
\tiny 
a) Milk collection dataset)
\begin{tabularx}{\textwidth}{X|c|c|c}
\hline
\textbf{Region} & \textbf{CAGE} & \textbf{SimpleAvg} & \textbf{wEnsemble} \\ \hline
Region A & \textbf{2.08} $\pm$ \textbf{1.88} & 4.41 $\pm$ 3.81 & 2.96 $\pm$ 2.31 \\ \hline
Region B & \textbf{2.90} $\pm$ 5.95 & 3.35 $\pm$ \textbf{5.79} & 3.77 $\pm$ 8.55 \\ \hline
Region C & \textbf{5.30} $\pm$ \textbf{2.57} & 6.49 $\pm$ 2.82 & 6.45 $\pm$ 2.65 \\ \hline
Region D & \textbf{1.31} $\pm$ \textbf{3.10} & 4.23 $\pm$ 7.99 & 4.07 $\pm$ 7.56 \\ \hline
Region E & \textbf{2.87} $\pm$ \textbf{3.26} & 4.93 $\pm$ 3.61 & 4.91 $\pm$ 3.92 \\ \hline
Region F & \textbf{1.83} $\pm$ 7.53 & 5.01 $\pm$ \textbf{7.47} & 6.61 $\pm$ 14.08 \\ \hline
\end{tabularx}
\end{minipage}%
\hfill
\begin{minipage}{0.49\textwidth}
\centering
\tiny 
b) owid-monkeypox dataset
\begin{tabularx}{\textwidth}{X|c|c|c}
\hline
\textbf{Country} & \textbf{CAGE} & \textbf{SimpleAvg} & \textbf{wEnsemble} \\ \hline
Australia & \textbf{0.035} $\pm$ \textbf{0.051} & 1.084 $\pm$ 0.282 & 0.147 $\pm$ 0.200 \\ \hline
Brazil & \textbf{0.192} $\pm$ 0.245 & 0.828 $\pm$ 0.290 & 0.184 $\pm$ \textbf{0.235} \\ \hline
Canada & \textbf{0.036} $\pm$ \textbf{0.043} & 1.093 $\pm$ 0.133 & 0.125 $\pm$ 0.210 \\ \hline
Japan & \textbf{1.129} $\pm$ \textbf{0.868} & 1.504 $\pm$ 1.125 & 1.421 $\pm$ 1.072 \\ \hline
South Africa & \textbf{0.000} $\pm$ \textbf{0.000} & 1.156 $\pm$ 0.403 & 0.576 $\pm$ 0.571 \\ \hline
United Kingdom & \textbf{0.019} $\pm$ \textbf{0.030} & 1.174 $\pm$ 0.104 & 0.090 $\pm$ 0.158 \\ \hline
\end{tabularx}
\end{minipage}
\end{table}

Table \ref{table:mape} details the MAE\% across different regions and countries, indicating the varied performance of the ensemble methods. In Region A, CAGE records the lowest mean MAE\% at 2.08, outperforming SimpleAvg at 4.41 and wEnsemble at 2.96. Similar patterns are observed in the owid-monkeypox dataset, where CAGE notably lowers MAE\% in Australia to 0.035 compared to SimpleAvg's 1.084 and wEnsemble's 0.147. 
These findings underline CAGE's effectiveness in enhancing forecasting accuracy, consistently outperforming other models in most scenarios. This consistent performance raises questions about the variability in data characteristics across regions and countries, potentially influencing the effectiveness of ensemble methods. Section \ref{stats} explores whether CAGE provides significant advantages over traditional methods.

\subsection{Statistical Significance test}
\label{stats}
In this section, we apply one-way ANOVA and posthoc t-tests for various ensemble methods across different regions and countries to determine the statistical significance of differences in performance between the methods. We follow the same steps as described in \cite{shahi2017dynamic,db2021classification}. The ANOVA test checks if there is any significant difference among the means of the three groups (CAGE, SimpleAvg, \& wEnsemble) overall. A significant ANOVA result (\textit{p}-value < 0.05) indicates that at least one of the group means is different from the others.

\begin{table}[h]
\centering
\caption{ANOVA tests of two datasets comparing MAE\% among ensemble methods}
\label{tab:anova_results}
\begin{minipage}{0.5\textwidth}
\centering
a) Milk Collection dataset
\rowcolors{2}{gray!15}{white} 
\begin{tabular}{c|c|c}
\hline
\textbf{Region} & \textbf{F-Statistic} & \textbf{\textit{p}-value} \\ \hline
Region A & 8.674 & \textbf{0.000} \\ 
Region B & 0.196 & 0.822 \\ 
Region C & 3.131 & \textbf{0.047} \\ 
Region D & 3.034 & 0.051 \\ 
Region E & 5.292 & \textbf{0.006} \\ 
Region F & 2.747 & 0.068 \\ \hline
\end{tabular}
\end{minipage}%
\begin{minipage}{0.5\textwidth}
\centering
b) owid-monkeypox dataset
\rowcolors{2}{gray!15}{white} 
\begin{tabular}{c|c|c}
\hline
\textbf{Country} & \textbf{F-Statistic} & \textbf{\textit{p}-value} \\ \hline
Australia & 57.08 & \textbf{0.000} \\ 
Brazil & 14.40 & \textbf{0.000} \\ 
Canada & 113.46 & \textbf{0.000} \\ 
Japan & 0.26 & 0.776 \\ 
South Africa & 14.36 & \textbf{0.000} \\ 
United Kingdom & 240.15 & \textbf{0.000} \\ \hline
\end{tabular}
\end{minipage}
\end{table}

Table \ref{tab:anova_results} presents ANOVA tests for both the milk collection and owid-monkeypox datasets, assessing significant differences in MAE\% among the ensemble methods (CAGE, SimpleAvg, \& wEnsemble). In the milk collection dataset, significant differences are noted in Regions A, C, and E with very low \textit{p}-values, highlighting variable efficacy among methods. Conversely, Regions B, D, and F show no significant differences, indicating similar performance across methods. For the owid-monkeypox dataset, countries like Australia, Brazil, Canada, and South Africa show significant differences with very low \textit{p}-values, suggesting one or more methods markedly outperform others. In contrast, Japan exhibits no significant differences, indicating comparable method performance. The UK shows substantial variability in method efficacy, as reflected by a very high F-statistic. These insights suggest varying levels of method suitability across different regions and countries, influenced possibly by local data characteristics or disease dynamics. Pairwise t-tests are illustrated in Table \ref{tab:ttest_pvalues} for further specifics on method comparisons, especially where significant differences are detected.

\begin{table}[h]
\centering
\caption{Posthoc t-test \textit{p}-values comparing MAE\% among ensemble methods across different datasets}
\label{tab:ttest_pvalues}
\resizebox{\textwidth}{!}{
\begin{tabular}{c|c|c|c}
\hline
\multicolumn{4}{|c|}{\textbf{Milk Collection Dataset}} \\ \hline
\textbf{Region} & \textbf{SimpleAvg vs CAGE} & \textbf{SimpleAvg vs wEnsemble} & \textbf{wEnsemble vs CAGE} \\ \hline
Region A & \textbf{0.000} & \textbf{0.025} & \textbf{0.041} \\ \hline
Region B & 0.701 & 0.779 & 0.560 \\ \hline
Region C & \textbf{0.031} & 0.950 & \textbf{0.031} \\ \hline
Region D & \textbf{0.019} & 0.918 & \textbf{0.020} \\ \hline
Region E & \textbf{0.004} & 0.978 & \textbf{0.006} \\ \hline
Region F & \textbf{0.040} & 0.489 & \textbf{0.041} \\ \hline
\multicolumn{4}{|c|}{\textbf{owid-monkeypox Dataset}} \\ \hline
\textbf{Country} & \textbf{SimpleAvg vs CAGE} & \textbf{SimpleAvg vs wEnsemble} & \textbf{wEnsemble vs CAGE} \\ \hline
Australia & \textbf{0.000} & 0.176 & \textbf{0.000} \\ \hline
Brazil & \textbf{0.001} & 0.949 & \textbf{0.001} \\ \hline
Canada & \textbf{0.000} & 0.295 & \textbf{0.000} \\ \hline
Japan & 0.499 & 0.586 & 0.890 \\ \hline
South Africa & \textbf{0.000} & \textbf{0.020} & \textbf{0.048} \\ \hline
United Kingdom & \textbf{0.000} & 0.269 & \textbf{0.000} \\ \hline
\end{tabular}
}
\end{table}

Posthoc t-test results for both the Milk Collection and owid-monkeypox datasets, as detailed in Table \ref{tab:ttest_pvalues}, show varying performances across different regions and countries. For the Milk Collection dataset, significant differences in MAE\% among all method pairs in Region A and selective significant differences in Regions C, D, E, and F highlight the CAGE method's superiority in most regions except Region B, where performances are similar across all methods. The lack of significant differences between SimpleAvg and wEnsemble across multiple regions suggests that these methods might be interchangeable under certain conditions. Conversely, for the owid-monkeypox dataset, significant differences in Australia, Brazil, Canada, South Africa, and the United Kingdom indicate distinct method performances, with CAGE often outperforming the others. The only exception is Japan, where results were mixed or non-significant, possibly due to unique regional characteristics or specific health interventions that may influence the effectiveness of the ensemble methods. These results demonstrate the effectiveness of CAGE and its robustness in various epidemiological and environmental contexts, suggesting its utility in settings where traditional methods may falter.

This differentiation by CAGE, primarily identified in the pairwise tests, underscores its effectiveness and the necessity for its use in settings where traditional methods may struggle.

\section{Conclusion}
In this paper, we proposed the Conformal Adversarial Generative Ensemble (CAGE), a novel forecasting methodology designed to address the limitations inherent in traditional ensemble techniques, particularly in handling extreme forecasts and noisy data. This innovative approach enhances forecast accuracy and provides a quantifiable measure of uncertainty, making it suitable for applications requiring robust decision-making support. Our empirical evaluations, using time series data on milk collection in various New Zealand regions and the public owid-monkeypox dataset detailing global health data, demonstrate CAGE’s superiority over traditional methods. These results suggest CAGE's broad applicability in diverse forecasting domains such as finance, healthcare, weather, and supply chain management, underscoring its potential as a powerful tool for robust decision-making in complex and dynamic environments.

\bibliographystyle{splncs04}
\bibliography{2155}

@inproceedings{dietterich2000ensemble,
  title={{Ensemble Methods in Machine Learning}},
  author={Dietterich, Thomas G},
  booktitle={International workshop on multiple classifier systems},
  pages={1--15},
  year={2000},
  doi = {10.1007/3-540-45014-9_1},
  organization={Springer}
}

@article{breiman1996bagging,
  title={{Bagging Predictors}},
  author={Breiman, Leo},
  journal={Machine learning},
  volume={24},
  pages={123--140},
  year={1996},
  doi = {10.1023/A:1018054314350},
  publisher={Springer}
}

@article{krawczyk2016learning,
  title={Learning from imbalanced data: open challenges and future directions},
  author={Krawczyk, Bartosz},
  journal={Progress in Artificial Intelligence},
  volume={5},
  number={4},
  pages={221--232},
  year={2016},
  doi = {10.1007/s13748-016-0094-0},
  publisher={Springer, Cham}
}

@article{liu2024ensemble,
  title={Ensemble Forecasts of Extreme Flood Events with Weather Forecasts, Land Surface Modeling and Deep Learning},
  author={Liu, Yuxiu and Yuan, Xing and Ji, Peng and Li, Chaoqun and An, Xindai},
  journal={Water},
  volume={16},
  number={7},
  pages={990},
  year={2024},
  doi={10.3390/w16070990},
  publisher={MDPI}
}

@article{shafer2008tutorial,
  title={{A Tutorial on Conformal Prediction}},
  author={Shafer, Glenn and Vovk, Vladimir},
  journal={Journal of Machine Learning Research},
  volume={9},
  number={3},
  pages = {371–421},
  doi = {10.5555/1390681.1390693},
  year={2008}
}

@book{vovk2005algorithmic,
  title={{Algorithmic Learning in a Random World}},
  author={Vovk, Vladimir and Gammerman, Alexander and Shafer, Glenn},
  year={2005},
  doi  = {10.1007/978-3-031-06649-8},
  publisher={Springer Science \& Business Media}
}

@book{vandeput2023demand,
  title={{Demand Forecasting Best Practices}},
  author={Vandeput, Nicolas},
  year={2023},
  publisher={Simon and Schuster}
}

@inproceedings{shahi2017dynamic,
  title={{Dynamic Real-Time Segmentation and Recognition of Activities Using a Multi-feature Windowing Approach}},
  author={Shahi, Ahmad and Woodford, Brendon J and Lin, Hanhe},
  booktitle={Trends and Applications in Knowledge Discovery and Data Mining: PAKDD 2017 Workshops, MLSDA, BDM, DM-BPM Jeju, South Korea, May 23, 2017, Revised Selected Papers 21},
  pages={26--38},
  year={2017},
  doi = {10.1007/978-3-319-67274-8_3},
  organization={Springer, Cham}
}

@article{db2021classification,
  title={Classification of oil palm female inflorescences anthesis stages using machine learning approaches},
  author={DB, Mamehgol Yousefi and Rafie, AS Mohd and Abd Aziz, Samsuzana and Azrad, Syaril and Masri, Mohamed Mazmira Mohd and Shahi, Ahmad and Marzuki, OF},
  journal={Information Processing in Agriculture},
  volume={8},
  number={4},
  pages={537--549},
  year={2021},
  doi={10.1016/j.inpa.2020.11.007},
  publisher={Elsevier}
}

@inproceedings{cheng2012comparison,
  title={A comparison of ensemble methods in financial market prediction},
  author={Cheng, Cheng and Xu, Wei and Wang, Jiajia},
  booktitle={2012 Fifth International Joint Conference on Computational Sciences and Optimization},
  pages={755--759},
  doi={10.1109/CSO.2012.171},
  year={2012},
  organization={IEEE}
}

@article{Sloughter2007ProbabilisticQP,
  title={Probabilistic Quantitative Precipitation Forecasting Using Bayesian Model Averaging},
  author={J. Mc Lean Sloughter and Adrian E. Raftery and Tilmann Gneiting and Chris Fraley},
  journal={Monthly Weather Review},
  volume={135},
  number={9},
  pages={3209-3220},
  year={2007},
  publisher={American Meteorological Society},
  doi={10.1175/MWR3441.1}
}

@article{He2023FinancialTS,
  title={Financial Time Series Forecasting with the Deep Learning Ensemble Model},
  author={Kaijian He and Qian Yang and Lei Ji and Jingcheng Pan and Yingchao Zou},
  journal={Mathematics},
  volume={11},
  number={4},
  pages={1054},
  year={2023},
  publisher={MDPI},
  doi={10.3390/math11041054}
}

@article{dutta2022early,
  title={Early prediction of diabetes using an ensemble of machine learning models},
  author={Dutta, Aishwariya and Hasan, Md Kamrul and Ahmad, Mohiuddin and Awal, Md Abdul and Islam, Md Akhtarul and Masud, Mehedi and Meshref, Hossam},
  journal={International Journal of Environmental Research and Public Health},
  volume={19},
  number={19},
  pages={12378},
  doi={10.3390/ijerph191912378},
  year={2022},
  publisher={MDPI}
}

@article{gupta2022stacking,
  title={Stacking ensemble-based intelligent machine learning model for predicting post-COVID-19 complications},
  author={Gupta, Aditya and Jain, Vibha and Singh, Amritpal},
  journal={New Generation Computing},
  volume={40},
  number={4},
  pages={987--1007},
  year={2022},
  publisher={Springer},
  doi={10.1007/s00354-021-00144-0}
}

@article{yoon2023multi,
  title={Multi-modal stacking ensemble for the diagnosis of cardiovascular diseases},
  author={Yoon, Taeyoung and Kang, Daesung},
  journal={Journal of Personalized Medicine},
  volume={13},
  number={2},
  pages={373},
  doi={10.3390/jpm13020373},
  year={2023},
  publisher={MDPI}
}

@article{bhagat2024efficient,
  title={An efficient stacking-based ensemble technique for early heart attack prediction},
  author={Bhagat, Monu and Sharma, Aayush and Agarwal, Piyanshi},
  journal={Multimedia Tools and Applications},
  pages={1--26},
  doi={10.1007/s11042-023-14625-5},
  year={2024},
  publisher={Springer}
}

@misc{duan2021comparing,
  title={Comparing Bayesian model averaging and reliability ensemble averaging in post-processing runoff projections under climate change. Water, 13 (15), 2124},
  author={Duan, K and Wang, X and Liu, B and Zhao, T and Chen, X},
  doi={10.3390/w13152124},
  year={2021}
}

@article{jakhar2024self,
  title={SELF: a stacked-based ensemble learning framework for breast cancer classification},
  author={Jakhar, Amit Kumar and Gupta, Aman and Singh, Mrityunjay},
  journal={Evolutionary Intelligence},
  volume={17},
  number={3},
  pages={1341--1356},
  year={2024},
  doi={10.1007/s12065-023-00824-4},
  publisher={Springer}
}

@inproceedings{adhikari2016time,
  title={Time series forecasting through a dynamic weighted ensemble approach},
  author={Adhikari, Ratnadip and Verma, Ghanshyam},
  booktitle={Proceedings of 3rd International Conference on Advanced Computing, Networking and Informatics: ICACNI 2015, Volume 1},
  pages={455--465},
  doi={10.1007/978-3-030-73103-8_43},
  year={2016},
  organization={Springer}
}

@phdthesis{Tang2023,
  author       = {Siyi Tang},
  title        = {Improving Deep Learning for Medical Time Series Data by Modeling Multidimensional Dependencies},
  school       = {Stanford University},
  year         = 2023,
  address      = {Stanford, California},
  note         = {Ph.D. dissertation}
}

@incollection{Yang2002Multistage,
  title={Multistage Neural Network Ensembles},
  author={Yang, Shuang and Browne, Antony and Picton, Philip D.},
  booktitle={Multiple Classifier Systems},
  editor={Roli, Fabio and Kittler, Josef},
  series={Lecture Notes in Computer Science},
  volume={2364},
  pages={87--96},
  year={2002},
  publisher={Springer},
  address={Berlin, Heidelberg},
  doi={10.1007/3-540-45428-4_9}
}

@article{brown2005managing,
  title={Managing diversity in regression ensembles},
  author={Brown, Gavin and Wyatt, Jeremy L and Tino, Peter},
  journal={Journal of Machine Learning Research},
  volume={6},
  pages={1621--1650},
  year={2005},
  publisher={JMLR.org},
  doi={10.5555/1046920.1088738}
}

@article{hammami2020neural,
  title={Neural networks for online learning of non-stationary data streams: a review and application for smart grids flexibility improvement},
  author={Hammami, Zeineb and Sayed-Mouchaweh, Moamar and Mouelhi, Wiem and Ben Said, Lamjed},
  journal={Artificial Intelligence Review},
  volume={53},
  number={8},
  pages={6111--6154},
  year={2020},
  doi={10.1007/s10462-020-09844-3},
  publisher={Springer}
}

@article{rokach2010ensemble,
  title={Ensemble-based classifiers},
  author={Rokach, Lior},
  journal={Artificial Intelligence Review},
  volume={33},
  number={1-2},
  pages={1-39},
  year={2010},
  doi = {10.1007/s10462-009-9124-7},
  publisher={Springer, Cham}
}

@book{kuncheva2014combining,
  title={Combining Pattern Classifiers: Methods and Algorithms},
  author={Kuncheva, Ludmila I},
  year={2014},
  doi  = {10.1002/9781118914564},
  publisher={John Wiley \& Sons}
}

@article{wolpert1992stacked,
  title={Stacked generalization},
  author={Wolpert, David H},
  journal={Neural Networks},
  volume={5},
  number={2},
  pages={241-259},
  doi = {10.1016/S0893-6080(05)80023-1},
  year={1992},
  publisher={Elsevier}
}

@article{ganaie2022ensemble,
  title={Ensemble deep learning: A review},
  author={Ganaie, Mudasir A and Hu, Minghui and Malik, Ashwani Kumar and Tanveer, Muhammad and Suganthan, Ponnuthurai N},
  journal={Engineering Applications of Artificial Intelligence},
  volume={115},
  pages={105151},
  year={2022},
  doi={10.1016/j.engappai.2022.105151},
  publisher={Elsevier}
}

@article{hu2021model,
  title={Model complexity of deep learning: A survey},
  author={Hu, Xia and Chu, Lingyang and Pei, Jian and Liu, Weiqing and Bian, Jiang},
  journal={Knowledge and Information Systems},
  volume={63},
  pages={2585--2619},
  year={2021},
  publisher={Springer},
  doi={10.1007/s10115-021-01605-0}
}

\end{document}